\documentclass[letterpaper]{article} 
\usepackage[submission]{aaai24}  
\usepackage{times}  
\usepackage{helvet}  
\usepackage{courier}  
\usepackage[hyphens]{url}  
\usepackage{graphicx} 
\usepackage{natbib}  
\usepackage{caption} 
\usepackage{algorithm}
\usepackage{algorithmic}

\usepackage{newfloat}
\usepackage{listings}
\DeclareCaptionStyle{ruled}{labelfont=normalfont,labelsep=colon,strut=off} 
\floatstyle{ruled}
\newfloat{listing}{tb}{lst}{}
\floatname{listing}{Listing}
\usepackage{graphicx}
\usepackage{mathtools}
\usepackage{tabularx}
\graphicspath{{pic/}}
\usepackage{enumitem}
\usepackage{booktabs}
\usepackage{amssymb}
\usepackage{caption}
\usepackage{subcaption}
\usepackage[dvipsnames]{xcolor}
\usepackage{lineno}
\usepackage[flushleft]{threeparttable}
\usepackage{multirow}

\title{A Cognitively-Inspired Neural Architecture for Visual Abstract Reasoning Using Contrastive Perceptual and Conceptual Processing}

\author {
    Yuan Yang,
    Deepayan Sanyal,
    James Ainooson,
    Joel Michelson,
    Effat Farhana,
    Maithilee Kunda\thanks{Corresponding Author.}
}
\affiliations {
    Vanderbilt University, Nashville TN 37240, USA\\
    yuan.yang@vanderbilt.edu, deepayan.sanyal@vanderbilt.edu, james.ainooson@vanderbilt.edu, joel.p.michelson@vanderbilt.edu, effat.farhana@vanderbilt.edu, mkunda@vanderbilt.edu,
}

\usepackage{bibentry}

\begin{document}

\maketitle

\begin{abstract}
We introduce a new neural architecture for solving visual abstract reasoning tasks inspired by human cognition, specifically by observations that human abstract reasoning often interleaves perceptual and conceptual processing as part of a flexible, iterative, and dynamic cognitive process. Inspired by this principle, our architecture models visual abstract reasoning as an iterative, self-contrasting learning process that pursues consistency between perceptual and conceptual processing of visual stimuli. We explain how this new Contrastive Perceptual-Conceptual Network (CPCNet) works using matrix reasoning problems in the style of the well-known Raven's Progressive Matrices intelligence test. Experiments on the machine learning datasets, the RAVEN family and PGM, show that CPCNet achieves higher accuracy than all previously published models while also using the weakest inductive bias. We also point out a substantial and previously unremarked class imbalance in the original RAVEN dataset, and we propose a new variant of RAVEN---AB-RAVEN---that is more balanced in terms of abstract concepts.  
\end{abstract}


\section{Introduction}

Analogy-making---the process of comparing and contrasting two or more things to enable additional relational inferences of various kinds---has been argued to be one of the foundational aspects of human intelligence \cite{hofstadter2013surfaces}.  So, how do humans make analogies?  

Consider the simple analogy in Figure \ref{fig:analogy}.  What relationships do you notice?  Initially, one might recognize fish and birds as animals that move around in the water and air, respectively.  Fish and birds both have similar body structures in terms of their heads, fins/wings, and tails.  However, one might further reflect that birds get propulsion from their wings, whereas many fish get propulsion from their tails.  This alternate mapping (bird wings to fish tails, and bird tails to fish fins) is influenced by conceptual processing of the initial perceptual inputs of the figure, and can in turn influence further perceptual processing of which similarities we emphasize and how we build our analogical representations.

Theories of human perceptual and conceptual systems \cite[e.g.,][]{barsalou1999perceptual}, including in the context of analogy-making \cite[e.g.,][]{carpenter1990one}, have made observations about this kind of bidirectional interplay between perceptual and conceptual processing, and forms of this interplay have also been explored in knowledge-based (i.e., symbolic) computational models of analogical reasoning \cite{lovett2017modeling}.  In this paper:

\begin{figure}[t]
    \centering
    \includegraphics[width=0.7\linewidth]{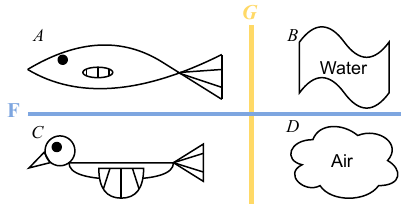}
    \caption{A is to B as C is to D.  But in what ways?}
    \label{fig:analogy}
\end{figure}

\begin{figure}[t]
    \centering
    \includegraphics[width=0.6\linewidth]{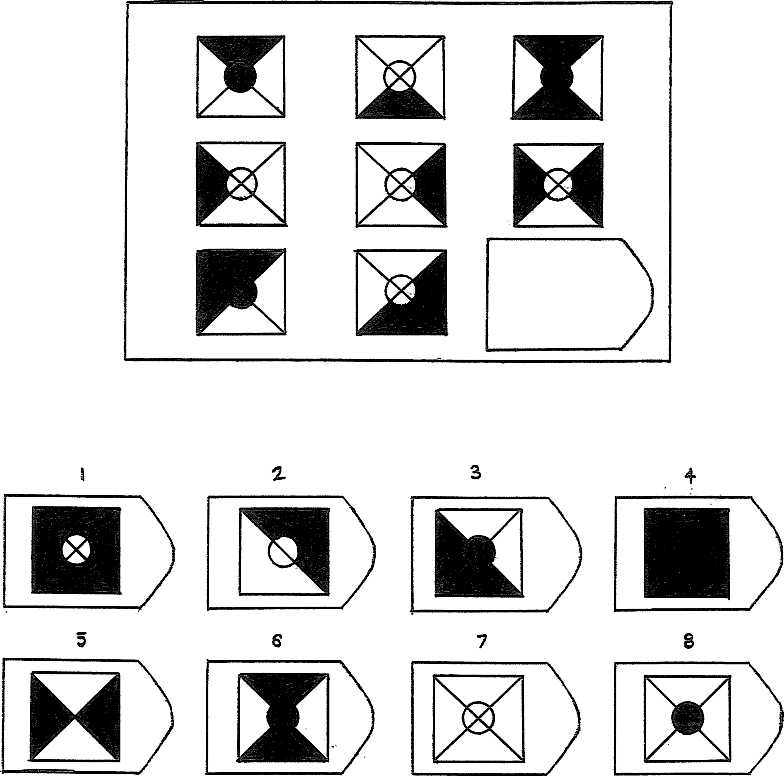}
    \caption{An example item of Raven's Progressive Matrices \cite{kunda2013computational}.}
    \label{fig:example-raven}
\end{figure}

\begin{enumerate}[nolistsep,noitemsep]
    \item We propose a new, cognitively-inspired Contrastive Perceptual-Conceptual neural Network (CPCNet) that models this kind of interplay between perceptual and conceptual processing in the context of visual abstract reasoning tasks like the example shown in Figure \ref{fig:example-raven}.
    \item Using the abstract reasoning datasets--RAVEN, I-RAVEN, RAVEN-FAIR, and PGM, we experimentally demonstrate that CPCNet is more effective than previous architectures by achieving the highest accuracy with the weakest inductive bias. 
    \item Finally, we point out a substantial, previously unremarked class imbalance in the original RAVEN dataset, and we propose a new variant---AB-RAVEN---that is more balanced in terms of abstract concepts.
\end{enumerate}

\section{Approaches to Visual Abstract Reasoning}

Raven's Progressive Matrices (RPM) is a family of human intelligence tests created by \citet{raven1936mental} about a century ago. 
RPM is acclaimed as the best single-format intelligence test that exists to date for evaluating the core intelligence factors, such as general intelligence and fluid intelligence \cite{snow1984topography,hunt2010human}. 

RPM is a kind of visual abstract reasoning task, where human subjects are expected to discover abstract patterns and concepts from raw visual stimuli, and apply these abstract patterns to reason about the visual stimuli \cite{raven2008raven}. Figure~\ref{fig:example-raven} gives an example item of RPM. It consists of a matrix of images with the last entry missing and multiple (usually eight) answer choices. To solve such an item, the human subject needs to select an answer choice to complete the matrix so that the abstract patterns among rows and columns are consistent.   For example, the abstract pattern in Figure~\ref{fig:example-raven} is that taking the union of the first two entries in a row (or a column) leads to the third entry in the row (or column), which leads to the correct answer of the fourth choice. 

\begin{figure}[!htbp]
     \centering
     \begin{subfigure}[b]{\linewidth}
         \centering
         \includegraphics[width=0.6\linewidth]{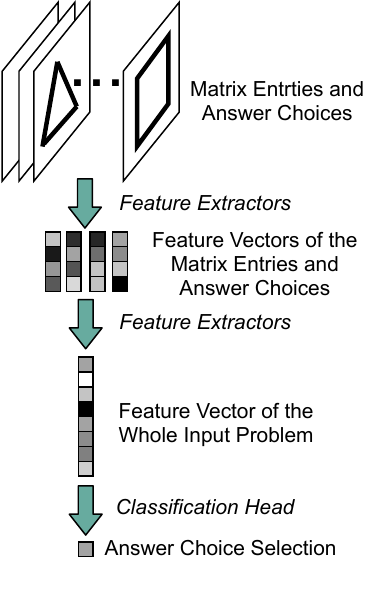}
         \caption{Image Classification Paradigm for Solving RPM}
         \label{fig:ic-rpm}
     \end{subfigure}

     \begin{subfigure}[b]{\linewidth}
         \centering
         \includegraphics[width=0.9\linewidth]{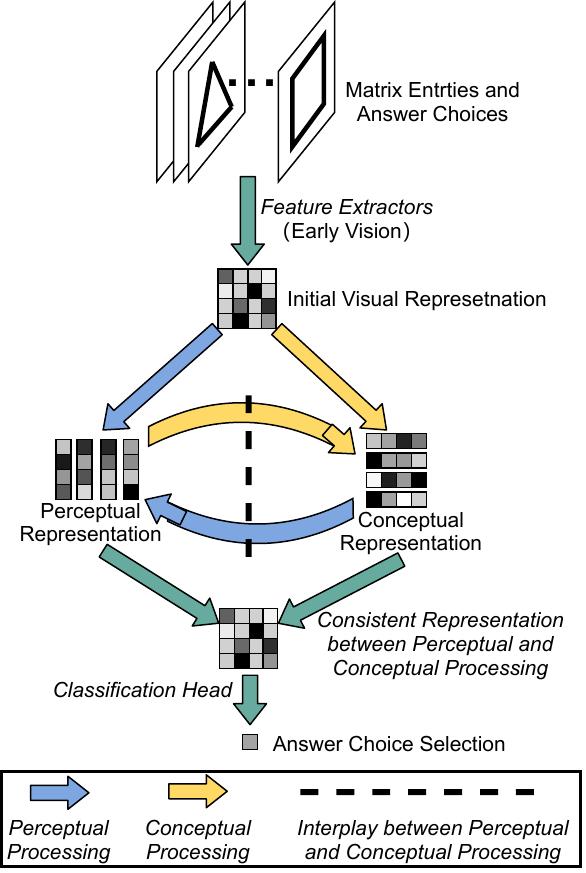}
         \caption{A Paradigm Simulating Human Cognition for Solving RPM}
         \label{fig:hc-rpm}
     \end{subfigure}

        \caption{Two Paradigms for Solving RPM. Note that the sizes and numbers of tensors are diagrammatic, not representing the implementation.}
        \label{fig:three graphs}
\end{figure}


In the recent surge of work using deep learning to tackle visual abstract reasoning, most deep neural network models have followed a standard image classification paradigm, as shown in Figure~\ref{fig:ic-rpm}.  Taking as input the raster images of matrix entries and answer choices, this paradigm repeatedly applies feature extractions as in the famous ImageNet work \cite{alex2012imagenet}, decreasing the sizes of the spatial dimensions but increasing the size of the channel dimension, until a single vector can be obtained to represent the entire input problem, and then a MLP classification head is appended to predict the class label, which is the index of the correct answer choice. 

An alternate approach leverages the observations about human cognition outlined in the introduction above, i.e., that reasoning can often be enhanced by interleaving perceptual and conceptual processing, allowing each process to influence the other.  Figure~\ref{fig:hc-rpm} illustrates this kind of approach. Taking the same raw visual stimuli as in the image classification paradigm, this alternate paradigm uses feature extractors, simulating early vision processing, to form an initial visual representation of input images. Then there follows two types of processing: (1) perceptual processing that refines the perceptual (visual) representation of input images, for example, refining blurry feature maps of lines and angles to clear feature maps of shapes, and (2) conceptual processing that enriches the representation of abstract concepts, i.e., the relations between input images.

Then comes the main difference between the image classification paradigm and this paradigm---these two types of processing form a dynamic cycle, in which the perceptual and conceptual processing depend on each other's output. This cycle allows for direct interplay between perceptual and conceptual processing. The cycle keeps running for multiple steps until a consistency between perceptual and conceptual processing is reached (thus adding a computational requirement for checking or representing the consistency at every step).  The resulting consistent representation takes on a dual role as perceptual and conceptual representation both and is used to predict the answer label. 

Figure~\ref{fig:hc-rpm} depicts reasoning on RPM-like problems as a complex, flexible, and dynamic process. While it is not difficult to mechanically construct a deep neural net that mimics this kind of processing, the real technical difficulty lies is how to optimize it properly------given its dynamics, how can we make sure the network steadily converges to achieve the consistency needed to drive robust reasoning?  We describe our solution to this challenge in this paper.

\section{Detailed Motivation}
We focus on the interplay between perceptual and conceptual representation, not only because we expect it to provide added flexibility and dynamics to a reasoning process, but also because this kind of entangled representation is frequently implied in human cognitive studies \cite{barsalou1999perceptual}. This theory of representation could also  be corroborated by the eye tracking studies on RPM \cite{carpenter1990one}, in which the subject's attention moved back and forth between the matrix entries, visiting each entry multiple times, rather than scanning the entries linearly (though other explanations also exist for such gaze phenomena). 

This section will explain the feasibility of our approach in terms of implementation and the rationale in terms of analogical reasoning. The effectiveness of this paradigm will be shown in the experiment section.

\subsection{Feasibility for implementation}

\begin{figure*}[!htbp]
    \centering
    \includegraphics[width=\linewidth]{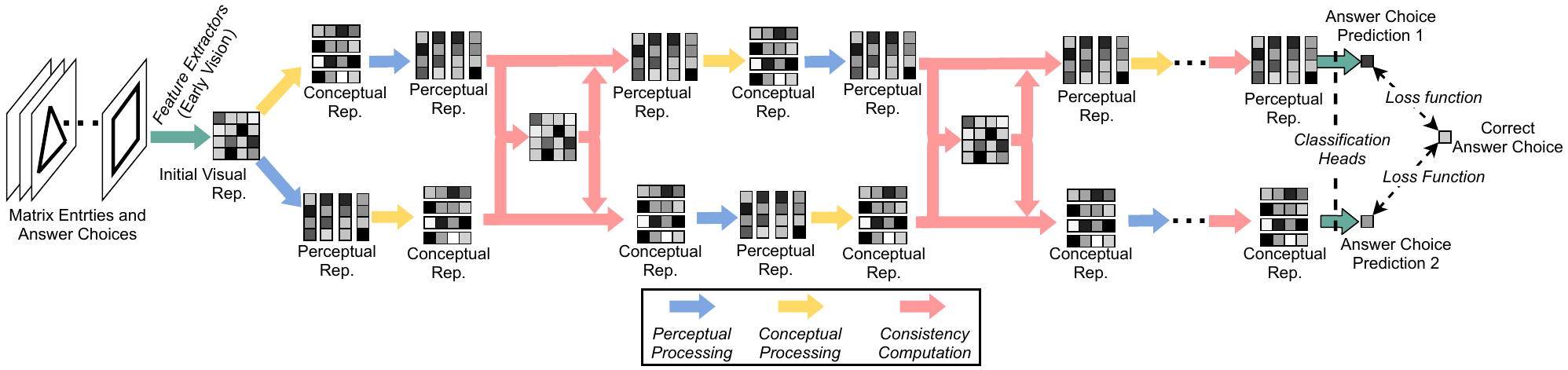}
    \caption{CPCNet: A Feed-Forward Architecture that unrolls the paradigm in Figure~\ref{fig:hc-rpm}}
    \label{fig:ff}
\end{figure*}

This subsection explains how this paradigm can be implemented and the implementation can behave as we described in the introduction section. Given the complex and dynamic nature of the cognitively-inspired paradigm, it is apparently inadvisable to mechanically compose multiple neural net modules into a neural net according to Figure~\ref{fig:hc-rpm}. Also, it is possible that the dynamic nature of the interplay could stop the training and inference from converging, if conditions that encourage convergence are not found.

In the feed-forward architectures commonly used in deep neural nets, we do not have this type of convergence issue. Thus, we can try to approximate the dynamic process with a  feed-forward network. Observe that there are two paths in Figure~\ref{fig:hc-rpm} that give the paradigm its dynamic nature, i.e., changing its own internal states---(1) \textit{Path 1} starting from the perceptual representation through the conceptual representation and returning back to the perceptual representation, and similarly, (2) \textit{Path 2} starting from the conceptual representation through the perceptual representation and returning back to the conceptual representation. Therefore, if we unroll the cycles with these two paths and add the consistency computation mentioned above, we will have a feed-forward architecture, as shown in Figure~\ref{fig:ff},  which \textbf{approximates} the fully iterative, cognitively-inspired paradigm in a feed-forward manner. We thus name our approach the Contrastive Perceptual-Conceptual Network (CPCNet). 

As indicated in the introduction section, this paradigm pursues consistency between perceptual and conceptual representations (the rationale will be explained later). There are two designs in the feed-forward architecture to make this happen. 

First, after each iteration of \textit{Path 1} and \textit{Path 2}, the red paths in Figure~\ref{fig:ff} first compute the consistency information between the perceptual and conceptual representations. This consistency information could be computed as a shared component between the perceptual and conceptual representations through a linear mapping, or more complex non-linear operations  could be used to have more representational power. Either way, the consistency information is then used to update the perceptual and conceptual representations, for example, deducting the consistency information from the perceptual and conceptual representations. Intuitively, this way, it would become easier and easier for later iterations to research a ``full'' consistency because the job of finding consistency is amortized over multiple iterations.

Second, the above computation structure only makes the consistency more likely to happen. But it does not necessarily happen. Thus, we designed two classification heads at the end of the architecture, which classify the perceptual and conceptual representations, respectively. Then, during training, the loss function is used to pull predictions of the perceptual and conceptual representations toward the same correct answer label. If the classification heads are not too deep, the pressure of loss function will go back through the classification heads and pull the perceptual and conceptual representations toward a consistent position. Here, the meaning of ``consistent'' becomes more clear---consistent representations could be mapped to the same correct answer label though some simple mappings, like a two-layer MLP. This design here is very similar to the idea of supervised contrastive learning \cite{Khosla2020supervised}, but it does not require data augmentation or pre-training. Instead, it relies on the delicate architecture design, which is inspired by the interleaved human cognitive process, to achieve the contrastive effect.

\subsection{Rationale for analogical reasoning}

Visual abstract reasoning tasks like the RPM can be considered as analogy tasks, because when an RPM item is solved, the human subject is making analogies between rows or columns (or both). To explain the rationale more clearly, let's consider the simpler visual analogy \textit{``A is to B as C is to D''} in Figure~\ref{fig:analogy} from the introduction in more depth. 

Suppose that a human subject has formed an initial visual representation for each analog in the analogy by looking at the figure for two seconds, but it is probably not the final, correct representation. 

According to the influential structure-mapping theory of analogy \cite{gentner1983structure}, the subject needs to construct a mapping $F$ between the base domain $(A, B)$ and the target domain$(C, D)$.  This mapping depends on how the analogs are represented. Given the initial visual representations of analogs, the fish and the bird are probably mapped to each other according to their appearance, e.g., \textit{head to head, fins to wings, and tail to tail}, and the air and the water are mapped  in a holistic way. 

Then, if the subject's thinking moves to a higher level and tries to map the relations (i.e., $G$ in Figure~\ref{fig:analogy})  in $(A, B)$ to the ones in $(C, D)$, she will find that they do not exactly match. In particular, many fish use tails for propulsion and fins for direction, whereas birds use wings for propulsion and tails for direction. This observation on $G$ updates the mapping $F$ and the representations of analogs---\textit{fish fins to bird tails, fish tails to bird wings, fish heads to bird heads}, and \textit{air to water holistically}. 

Given this clearer mapping $F$, if the subject moves up to a higher level again and compare the relations $G$, the mapping between $B$ and $D$ could be further refined to \textit{air dynamics is mapped to fluid dynamics} (rather than their colors) and thus the representation of water and air are also updated to focus on their dynamics properties.

If the subject can give initial representations of analogs that can directly lead to the final correct mappings $F$ and relations $G$, she may not need to go through any iterative process like this. However, in real-life situations where stimuli have so many meanings and connections, the correct representations of analogs cannot always be formed immediately. This iterative process of working on $F$, $G$, and the representations of analogs is always needed to make and understand analogies. Given $F$ corresponding to the perceptual processing and $G$ corresponding to the conceptual processing, this iterative process is equivalent to the interplay between perceptual and conceptual processing.

About the desire for consistency, its rationale follows from an assumption that the analogy is completely interpreted or understood only if the iterative process has ended, i.e., no updates are being made to representations of analogs anymore. 

In other words, it has been well recognized that analogical proportions enjoy central permutation as a characteristic property \cite{dorolle1949raisonnement}. That is, \textit{A is to B as C is to D} if and only if \textit{A is to C as B is to D}. This corresponds to interpretations of the analogy in Figure~\ref{fig:analogy} in the horizontal or vertical direction. Two directions are equivalent. That one direction holds implies that the other direction also holds.  Given this symmetry, $G$ could also be regarded as a mapping between $(A,C)$ and $(B, C)$. If the interpretation of the analogy is unique, i.e., the mappings are unique, we will have $F \circ G = G \circ F$, i.e., $F$ and $G$ are commutative. 

This equation is a very concise and beautiful description of analogy-making. And by pursuing consistency between perceptual and conceptual processing, we are actually pursuing equality in this equation, albeit in a data-driven and approximate way.

\section{Related Work}

There is a long line of research on computational solutions to RPM tasks, especially RPM-like datasets. Reviewing every one of them would be impossible here. We thus use a taxonomy to classify them into four categories and briefly describe each of them. More extensive reviews can be found in \citep{yang2022conceptual,malkinski2022deep,malkinski2022review}.

\textbf{Imagery-Based Approach.}  
Visual mental imagery refers to mental imagistic representation in human cognition \citep{kosslyn2006case}. It plays a crucial role in human visual reasoning ability. The most important characteristic of mental imagery is that human can experience mental imagery in the absence of the concurrent sensory input. The imagery-based approach simulates human mental imagery by directly operating on the raw visual input and, through mental operations like rotation, addition, and subtraction, it can solve a substantial portion of original RPM tests \citep{yang2020not,mcgreggor2014fractals}.

\textbf{Logical Reasoning.}  
The computational models using logical reasoning work on symbolic representations of RPM items and reason in formal logic. For example, an entry image $A$ in a matrix could be described by a set of propositions: ``triangle($A$)=True, triangle-large($A$)=False, triangle-on-the-left($A$)=True, square($A$)=True, square-small($A$)=True, etc''. The symbolic representations in these models are manually constructed or obtained through a preprocessing module. Representatives of this approach are ANALOGY \citep{evans1964program}, FAIRMAN ,and BETTERMAN \citep{carpenter1990one}.

\textbf{Neuro-Symbolic Approach.}  
The neuro-symbolic models consist of two modules---a neural perception frontend and a symbolic reasoning backend. The neural perception frontend (usually implemented as neural nets) extracts the symbolic representation of entry images (including but not limited to logical representation and probability representation) which are based on a predefined formal representation system. The symbolic reasoning backend performs symbolic manipulation or probability calculation according to the predefined formal representation system. Neuro-symbolic approach and the next approach---learning models---are data-driven approach, whereas the first two approaches are knowledge-based approaches. Examples of neuro-symbolic models for solving RPM-like tasks include ALANS2, PrAE, VAE-GPP, TRIVR, LoGe, NVSA and AMR\citep{zhang2022learning, zhang2021abstract, shi2021raven, he2021two, yu2021abstract, hersche2023neuro, xu2023abstract}.  

\textbf{Learning Models.}  
Unlike the previous approach, learning approach does not rely on any predefined representation system of geometric objects and abstract patterns. Instead, the representations are learned from raw perceptual input and represented as feature vectors. When this paper is written, almost all of the popular deep learning architectures have been experimented on RPM tasks, such as CNN, ResNet family, recurrent neural networks, and attention models \cite{hu2021stratified, benny2021scale, sahu2023savir, wei2023multi}. This approach has become more and more popular recently because of large RPM-like datasets created in the last five years \citep{barrett2018measuring,zhang2019raven}. 

However, it needs to be pointed out that these datasets are not perfect. For example, the RAVEN dataset \cite{zhang2019raven} is flawed because of the context-blind issue, i.e., training only on the answer choices leads to good performance\cite{hu2021stratified}. Thus, two variants of RAVEN---I-RAVEN \cite{hu2021stratified} and RAVEN-FAIR\cite{benny2021scale}---were proposed to fix this issue. Besides different variants of RAVEN, the evaluation setting of learning models is a more complicated issue. We will elaborate more on this in the experiment section. 


\section{CPCNet for Solving RPM-Like Datasets}

\begin{figure*}[!htbp]
    \centering
    \includegraphics[width=0.8\linewidth]{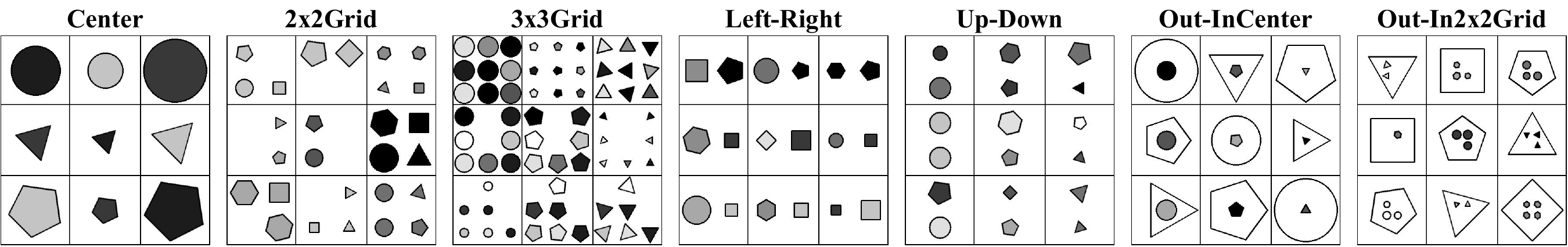}
    \caption{7 spatial configurations of the RAVEN dataset. Each configuration is illustrated by a complete 3x3 matrix. In the \textbf{Center} configuration, each entry contains only one object located at the center. In \textbf{2x2Grid} and \textbf{3x3Grid}, objects can only be located on the grid positions in each entry. In \textbf{Left-Right} and \textbf{Up-Down}, each entry contains exactly two objects located at the fixed positions shown in the figure. \textbf{Out-InCenter} is a combination of two center configurations and \textbf{Out-InGrid} is a combination of a center configuration and a 2x2Grid configuration.}
    \label{fig:raven_configs}
\end{figure*}

\begin{table*}[!htbp]
\centering
\begin{tabular}{ | c | c | c | c | c | c | c | }
\hline
 & Number	& Position & Number/Position &	Type & Size & Color  \\ \hline
Constant	                & 0 & 0                                                               & \textcolor{OliveGreen}{31803} & \textcolor{OliveGreen}{19240} &	\textcolor{OliveGreen}{16451} &	\textcolor{OliveGreen}{21817} \\ \hline
Progression	            & \textcolor{RawSienna}{1857} &	\textcolor{RawSienna}{1809} &	0	& \textcolor{OliveGreen}{19386} &	\textcolor{OliveGreen}{14049} &	\textcolor{OliveGreen}{12797} \\ \hline
Arithmetic              & \textcolor{RawSienna}{872} & \textcolor{RawSienna}{2933} &	0 &	0 &	\textcolor{OliveGreen}{13012} &	\textcolor{OliveGreen}{12686} \\ \hline
Distribute-Three     & \textcolor{RawSienna}{2925} & \textcolor{RawSienna}{2791} &	0	& \textcolor{OliveGreen}{19447}	&\textcolor{OliveGreen} {16355} &	\textcolor{OliveGreen}{12767} \\ \hline
Sum by Color & \multicolumn{2}{r}{\textcolor{RawSienna}{15187}} & \multicolumn{4}{|r|}{\textcolor{OliveGreen}{209810}}  \\ \hline
\end{tabular}
\caption{Numbers of RAVEN (I-RAVEN and RAVEN-FAIR) items containing each combination of abstract rules and attributes.}
\label{tab:raven-count}
\end{table*}

\begin{table*}[!htbp]
\centering
\begin{tabular}{ | c | c | c | c | c | c | c | }
\hline
 & Number	& Position & Number/Position &	Type & Size & Color  \\ \hline
Constant	                & 0 & 0                                                               & \textcolor{OliveGreen}{19574} & \textcolor{OliveGreen}{17507} &	\textcolor{OliveGreen}{15263} &	\textcolor{OliveGreen}{21220} \\ \hline
Progression	            & \textcolor{RawSienna}{4058} &	\textcolor{RawSienna}{4040} &	0	& \textcolor{OliveGreen}{17641} &	\textcolor{OliveGreen}{11998} &	\textcolor{OliveGreen}{11010} \\ \hline
Arithmetic              & \textcolor{RawSienna}{6329} & \textcolor{RawSienna}{6307} &	0 &	0 &	\textcolor{OliveGreen}{11508} &	\textcolor{OliveGreen}{10932} \\ \hline
Distribute-Three     & \textcolor{RawSienna}{6335} & \textcolor{RawSienna}{6421} &	0	& \textcolor{OliveGreen}{17579}	&\textcolor{OliveGreen} {15080} &	\textcolor{OliveGreen}{10907} \\ \hline
Sum by Color & \multicolumn{2}{r}{\textcolor{RawSienna}{33490}} & \multicolumn{4}{|r|}{\textcolor{OliveGreen}{180219}}  \\ \hline
\end{tabular}
\caption{Numbers of AB-RAVEN items containing each combination of abstract rules and attributes.}
\label{tab:ab-raven-count}
\end{table*}

Based on the above discussion about the feasibility and rationale, we can now formalize our method for solving RPM-like problems. In the main paper, we describe what kind of operations are applied at each step; detailed implementation and hyper-parameters can be found in the supplementary material.

We use the single-choice evaluation protocol, which is more challenging than the commonly-used multi-choice evaluation protocol, because comparing answer choices gives the model advantage over evaluating each single answer choice independently (see more about this in \cite{benny2021scale}). Thus, by inserting each answer choice into the matrix and evaluating them individually, we actually turn every multi-choice item into eight binary classification items, where the input to our model is a real tensor $x$ of shape $(R=rows, C=columns, H_{orig}=height, W_{orig}=width, channels=1)$ and the class label $y$ indicates whether the inserted answer choice is correct. 

For the feature extractor that simulates early vision in Figure~\ref{fig:ff}, we adopt a convolution-based encoder $f_E$ whose output channel number is set to $K > 1$.  Since the early vision usually does not involve forming abstract relations which are between entry images, $f_E$ is applied to encode each matrix entry individually:
\begin{align}
z_{r,c} &= f_E ( x_{r, c} ) \: \forall (r,c) \in \{ 1, \dots, R \} \times \{ 1, \dots, C \} \\
z &= [z_{r, c}]_{r=1, \dots, R, c=1, \dots, C} \in \mathbb{R}^{R \times C \times H \times W \times K }
\end{align}
where $H < H_{orig}$ and $W < W_{orig}$ as channels are increased from 1 to $K$. Let $z^{(0)}_1 = z^{(0)}_2 = z$ for the following Path 1 and 2, respectively.

For each iteration $i \in \{ 1, \dots, L \}$ after $f_E$, we need to define Path 1, Path 2, and the consistency computation between them.  For Path 1, we define perceptual and conceptual processing as convolution-based modules $h^{(i)}_1$ and $g^{(i)}_1$, respectively.  Similarly, for Path 2, we define the perceptual and conceptual processing as convolution-based modules $h^{(i)}_2$ and $g^{(i)}_2$.  For the consistency computation, we define a two-layer MLP $q^{(i)}$. The hyper-parameters of these modules are all set to values that preserve the input tensor shape $(R, C, H, W, K)$, i.e., the output channels of $h^{(i)}_1$, $g^{(i)}_1$,  $h^{(i)}_2$, and $g^{(i)}_2$ and the output units of $q^{(i)}$ are all set to $K$. 

For RPM task, the abstract concepts lie in the row and column dimensions as the abstract concepts are represented by rows and columns. We thus apply the convolutions of conceptual processing $g^{(i)}_1$ and $g^{(i)}_2$ on the $(R, C)$ dimensions of the input tensor, and apply the convolutions of perceptual processing $h^{(i)}_1$ and $h^{(i)}_2$ on  on the $(H, W)$ dimensions. And the consistency computation $q^{(i)}$ is applied on the channel dimension. Note that dimensions when not being computed are treated as transparent, i.e., like extended batch dimensions. Let the outputs from Path 1 and 2 of Iteration $i-1$ be $z^{(i-1)}_1$ and $z^{(i-1)}_2$, the computation of Iteration $i$ is:
\begin{align}
u_1 &= h^{(i)}_1 \circ g^{(i)}_1 (z^{(i-1)}_1) \\
u_2 &= g^{(i)}_2 \circ h^{(i)}_2 (z^{(i-1)}_2) \\
v_1 &= q^{(i)} (u_1) \\
v_2 &= q^{(i)} (u_2) \\
z^{(i)}_1 &= u_1 - v_2 \\
z^{(i)}_2 &= u_2 - v_1
\end{align}  

At last, we define two classification heads $p_1$ and $p_2$ for Path 1 and Path 2, respectively. 
\begin{align}
\hat{y}_1 &= p_1(flatten(mean(z^{(L)}_1))) \\
\hat{y}_2 &= p_2(flatten(mean(z^{(L)}_2)))
\end{align}
where the $mean$ takes the mean over the channel dimension of size $K$ and the $flatten$ flattens the input to a vector of length $R \times C \times H \times W$. 
For training, we compute binary cross entropy losses for both $\hat{y}_1$ and $\hat{y}_2$ with respect to $y$ and add them up as the final loss. For testing, we simply add up the $z^{(L)}_1$ and $z^{(L)}_2$ as a score of the input $x$ and select the highest score among all the answer choices.

\section{Experiments}

\begin{table*}[!htbp]
\centering
\begin{tabular}{l | l l l l l l l l l }
\toprule
     & Model             & Avg. Acc.   & Center          & 2x2Grid        & 3x3Grid        & L-R          & U-D            & O-IC           & O-IG \\
\midrule

\multirow{5}{2.2cm}{Multi-Choice Evaluation Protocol}
& LEN        & 78.3\%       & 82.3\%         & 58.5\%         & 64.3\%         & 87.0\%       & 85.5\%         & 88.9\%         & 81.9\%    \\
& MXGNet     & 83.91\%      & -              & -              & -              & -            & -              & -              & -         \\
& CoPINet    & 91.42\%      & 95.05\%        & 77.45\%        & 78.85\%        & 99.10\%      & 99.65\%        & 98.50\%        & 91.35\%   \\
& DCNet      & 93.58\%      & 97.80\%        & 81.70\%        & 86.65\%        & 99.75\%      & 99.75\%        & 98.95\%        & 91.45\%   \\
& SAVIR-T    & 94.0\%       & 97.8\%         & 94.7\%         & 83.8\%         & 97.8\%       & 98.2\%         & 97.6\%         & 88.0\%    \\ \hline

\multirow{11}{2.2cm}{Single-Choice Evaluation Protocol}
& WReN        & 14.69\%      & 13.09\%        & 28.62\%        & 28.27\%        & 7.49\%       & 6.34\%         & 8.38\%         & 10.56\%   \\
& ARNe        & 19.67\%      & -              & -              & -              & -            & -              & -              & -         \\ 
& NCD         & 39.66\%      & 45.45\%        & 35.50\%        & 39.50\%        & 34.85\%      & 33.40\%        & 40.25\%        & 30.00\%   \\ 
& PrAE        & 65.03\%      & 76.50\%        & 78.60\%        & 28.55\%        & 90.05\%      & 90.85\%        & 48.05\%        & 42.60\%   \\
& ALANS       & 74.4\%       & 69.1\%         & 80.2\%         & 75.0\%         & 72.2\%       & 73.3\%         & 76.3\%         & 74.9\%    \\
& MRNet       & 84.0\%       & -              & -              & -              & -            & -              & -              & -         \\
& NVSA        & 87.7\%       & 99.7\%         & 93.5\%         & 57.1\%         & 99.8\%       & 99.1\%         & 98.1\%         & 65.4\%    \\
& SCL         & 91.6\%       & 98.1\%         & 91.0\%         & 82.5\%         & 96.8\%       &  96.5\%        & 96.0\%         & 80.1\%     \\
& AlgebraicMR & 92.9\% & 98.8\%        & 91.9\%         & \textbf{93.1\%}& 99.2\%       & 99.1\%         & 98.2\%         & 70.1\%     \\ 
& Rel-AIR     & 94.1\%       & 99.0\%         & 92.4\%         & 87.1\%         & 98.7\%       & 97.9\%         & 98.0\%         & 85.3\%     \\
& CPCNet(Ours)  &\textbf{96.92\%}&\textbf{100.0\%}&\textbf{96.70\%}  & 86.05\% & \textbf{100.0\%} & \textbf{99.90\%} & \textbf{99.90\%} & \textbf{95.90\%}\\
\hline    
  &   Human    & 84.4         & 95.5\%         & 81.8\%         & 79.6\%         & 86.4\%       & 81.8\%         & 86.4\%         & 81.8\% \\
\bottomrule
\end{tabular}
\caption{Accuracies on the original RAVEN. We report without-auxiliary-training accuracies if possible. Data source for each row: \citep{zheng2019abstract}, \citep{wang2020abstract}, \citep{zhang2019learning}, \citep{zhuo2021effective}, \citep{sahu2023savir}, \citep{zhang2019raven}, \citep{hahne2019attention}, \citep{zhuo2021unsupervised}, \citep{zhang2021abstract}, \citep{zhang2022learning}, \citep{benny2021scale}, \citep{hersche2023neuro}, \citep{wu2021scattering}, \citep{xu2023abstract},  and \citep{spratley2020closer}.}
\label{tab:acc-raven}
\end{table*}

\begin{table*}[!htbp]
\centering
\begin{tabular}{ l l l l l l l l l }
\toprule
      Model               & Avg. Acc.     & Center         & 2x2Grid        & 3x3Grid         & L-R                 & U-D               & O-IC                & O-IG \\
\midrule
     CPCNet(ours)                   & 98.84\%      & 99.75\%       & 99.20\%       & 94.95\%         & 99.70\%         & 99.80\%        & 99.50\%           & 98.95\%    \\ 
\bottomrule
\end{tabular}
\caption{Accuracies on the AB-RAVEN, using single-choice evaluation protocol.}
\label{tab:acc-ab-raven}
\end{table*}

\subsection{Datasets and a Striking Observation}

We did our experiments on the RAVEN dataset \cite{zhang2019raven} and its variants. RAVEN items have 7 spatial configurations to organize the geometric objects in matrix entries (see Figure~\ref{fig:raven_configs}). Each configuration has 6000, 2000, and 2000 items for training, validation, and test, respectively. 

A very striking result from almost all previous works on RAVEN is that the accuracies on 2x2Grid, 3x3Grid, and Out-InGrid are always significantly lower than on other item configurations. Some argue this result is because specific abstract rules (i.e., concepts, relations, or patterns) are difficult to learn \cite{wei2023multi}; some argue that the noise attributes in the grid configurations causes this result \cite{xu2023abstract}. Although these arguments are not wrong, the fundamental reason for this result might be a much more mundane (but apparently previously unremarked!) one---that RAVEN is an extremely imbalanced dataset in terms of the abstract rules represented in its problems. 

This point can be seen in Table~\ref{tab:raven-count}, which counts dataset items for each combination of abstract rules and attributes. There are two types of combinations in this table---the red ones that exist only in the 2x2Grid, 3x3Grid, and Out-InGrid configurations and the green ones that exist mainly in the other four configurations and exist in roughly 65\% of the items of 2x2Grid, 3x3Grid, and Out-InGrid. Moreover, the sum of the green ones is roughly 15 times of the sum of the red ones. Therefore, the red ones are much less represented both in their own configurations and globally. This argument also applies to RAVEN's variants, such as I-RAVEN and RAVEN-FAIR because they share the same set of abstract rules and attributes with the the original RAVEN.

As we all know, deep learning models require sufficient data to work properly and every deep learning model has a lower limit for training data. We thus hypothesize that it is because the reds ones in Table~\ref{tab:raven-count} are less represented that previous models usually work relatively worse on 2x2Grid, 3x3Grid, and Out-InGrid. To verify that this, we constructed a new variant of RAVEN which is more \textbf{B}alanced in terms of \textbf{A}bstract rules and attributes (we thus call it AB-RAVEN). Table~\ref{tab:ab-raven-count} shows the statistics of AB-RAVEN. It was made more balanced by decreasing the number of non-grid training items and increasing the number of grid training items while keeping the overall size of training set unchanged. The validation and test sets of AB-RAVEN remain the same as RAVEN's. If the hypothesis is true, training on this dataset will lead to a smaller gap between grid and non-grid (testing) accuracies. Meanwhile, this dataset can also check if previous models' high accuracies on non-grid configurations are a result of excessively many training items of non-grid configurations.  As you can see in Table~\ref{tab:ab-raven-count}, AB-RAVEN is not perfectly balanced, but just just more balanced than RAVEN. This is because making it  perfectly balanced will violate the design of 7 configurations, i.e., need to remove all non-grid configuration items. More details about AB-RAVEN is provided in the supplementary material.

\subsection{Results and Discussion}

We designed our model to avoid using meta-targets and structure information for auxiliary training because this kind of handcrafted auxiliary information is not always available for human RPM tests and general visual abstract reasoning tasks; instead, only the score of each answer choice is predicted individually and the highest score is selected as the answer, i.e., using the single-choice evaluation protocol, which is more difficult than the opposite---multi-choice evaluation protocol \cite{benny2021scale}. 

Although not discussed very often in literature, it has been shown by multiple works \cite{wei2023multi, sahu2023savir, wu2021scattering, benny2021scale, xu2023abstract}  that when using single-choice evaluation, i.e., not allowing the model to comparing answer choices before scoring them and thus not allowing it to use the backdoor of the original RAVEN, the original RAVEN is more challenging than I-RAVEN and RAVEN-FAIR, that is, the same model always achieves a higher accuracy on I-RAVEN and RAVEN-FAIR than on RAVEN. 

This makes sense because  
the way the answer choices of RAVEN were generated makes the answer choices more similar to each other than in I-RAVEN and RAVEN-FAIR  and thus more confusing to the model when evaluated individually; on the contrary, the ways answer choices were generated in I-RAVEN and RAVEN-FAIR make the distractors differ from the correct answer by more attributes and thus less confusing to the model when evaluated individually. Therefore, due to the page limit, we report only the results on datasets of interest to us---RAVEN and AB-RAVEN---here. Other results can be found in the supplementary material.

Table~\ref{tab:acc-raven} shows that our model achieve the best average accuracy compared to previous models and the best configuration accuracies on 6 out of 7 configurations. 
Although our model's accuracy is only 2.82\%$\sim$2.92\% higher than the second and third highest accuracies of Rel-AIR and SAVIR-T, our model is solving the RAVEN dataset in a more difficult and general way---SAVIR-T uses the the easier multi-choice evaluation protocol and is designed to utilize the inductive bias that is specific to RPM \cite{sahu2023savir}, and while Rel-AIR uses the harder single-choice evaluation, Rel-AIR employs a separately-trained entry-encoder to explicitly extract values of size and position attributes, which are also specific to RPM rules. Many of the models in Table~\ref{tab:acc-raven} more or less used inductive bias that is specific to RAVEN either in the model design or in the training procedure. On the contrary, our inductive bias, if we consider it as a kind of inductive bias, is the interplay and consistency between perceptual and conceptual processing, which is more meaningful for solving and understanding general visual abstract reasoning. In particular, CoPINet and DCNet, which have been reported to utilize the backdoor of RAVEN \cite{hu2021stratified}, also achieved lower accuracies than ours.

Table~\ref{tab:acc-ab-raven} shows our model's accuracy on the AB-RAVEN dataset. Our model achieved a nearly perfect overall accuracy and, compared to Table~\ref{tab:acc-raven}, the accuracy gap between grid and non-grid configurations has been reduced from 7.07\% to 1.98\%. This verifies our hypothesis about the imbalanced dataset in the last subsection. 

Besides the RAVEN family, we also tested CPCNet on PGM and achieved the best accuracy of 98.4\%. More details can be found in the supplementary material.

\section{Conclusion}

In this paper, we investigated a cognitively-inspired paradigm for solving visual abstract reasoning tasks by leveraging the interplay between perceptual and conceptual processing.  We discussed feasibility of this approach and detailed rationale, and, we designed a deep neural net architecture to simulate the interplay of processes suggested in the cognitively-inspired paradigm, i.e., the interplay and consistency between perceptual and conceptual processing. Experiments shows that our CPCNet architecture is more effective than all previous models for solving RPM-like datasets. Moreover, with a balanced dataset (AB-Raven) in terms of abstract concepts, this architecture can even achieve a nearly perfect performance.

\bibliography{ref}





\onecolumn
\thispagestyle{plain}
\begin{center}
    \Large
    \textbf{Supplementary Material for A Cognitively-Inspired Neural Architecture for Visual Abstract Reasoning Using Contrastive Perceptual and Conceptual Processing}
    
        
       
\end{center}

\begin{abstract}
This document is a supplementary material for the paper \textit{A Cognitively-Inspired Neural Architecture for Visual Abstract Reasoning Using Contrastive Perceptual and Conceptual Processing}. In this document, we provide more information about (1) the new variant of RAVEN---the AB-RAVEN dataset, (2) implementation details about the CPCNet architecture used in our experiments, (3) more experimental results on I-RAVEN and RAVEN-FAIR, (4) ablation studies of CPCNet, and (5) Experiments on PGM.
\end{abstract}

\section{AB-RAVEN}

\begin{table*}[!h]
    \centering
    \begin{tabular}{| l | l | l | l | l | l | l | l | l | l|}
    \hline
                   &    &  \multicolumn{4}{c|}{RAVEN (I-RAVEN and RAVEN-FAIR)}    & \multicolumn{4}{c|}{AB-RAVEN} \\ \hline
                   & Configuration &  Training & Validation & Test  & All Splits            &  Training & Validation & Test & All Splits  \\ \hline
    \multirow{3}{2.2cm}{Grid Configurations}
           & 2x2Grid     &  6000 & 2000 & 2000 & 10000                                   & 12400 &  2000 & 2000 & 16400 \\   \cline{2-10}
           & 3x3Grid     &  6000 & 2000 & 2000 & 10000                                   & 12400 &  2000 & 2000 & 16400 \\   \cline{2-10}
           & O-IG        &  6000 & 2000 & 2000 & 10000                                   & 12400 &  2000 & 2000 & 16400 \\   \hline
    \multirow{4}{2.2cm}{Non-Grid Configurations}       
           & Center      &  6000 & 2000 & 2000 & 10000                                   & 1200  &  2000 & 2000 & 5200 \\    \cline{2-10}
           & L-R         &  6000 & 2000 & 2000 & 10000                                   & 1200  &  2000 & 2000 & 5200 \\    \cline{2-10}
           & U-D         &  6000 & 2000 & 2000 & 10000                                   & 1200  &  2000 & 2000 & 5200 \\    \cline{2-10}
           & O-IC        &  6000 & 2000 & 2000 & 10000                                   & 1200  &  2000 & 2000 & 5200 \\    \hline
      Sum  &             &  42000 & 14000 & 14000 & 70000                                & 42000 & 14000 & 14000 & 70000 \\  \hline   
    \end{tabular}
    \caption{The number of items of each spatial configuration in RAVEN (I-RAVEN and RAVEN-FAIR) and AB-RAVEN.}
    \label{tab:config-raven-ab-raven}
\end{table*}

\begin{table*}[!h]
\centering
\begin{tabular}{ | c | c | c | c | c | c | c | }
\hline
                   & Number	                  & Position                    & Number/Position               & Type                          & Size                          & Color                         \\ \hline
Constant	       & 0                           & 0                           & \textcolor{OliveGreen}{6000}  & \textcolor{OliveGreen}{1977}  & \textcolor{OliveGreen}{1520} & \textcolor{OliveGreen}{1524}   \\ \hline
Progression	       & \textcolor{RawSienna}{0}    & \textcolor{RawSienna}{0}    & 0                             & \textcolor{OliveGreen}{1998}  & \textcolor{OliveGreen}{1465} & \textcolor{OliveGreen}{1454}   \\ \hline
Arithmetic         & \textcolor{RawSienna}{0}    & \textcolor{RawSienna}{0}    & 0                             & 0                             & \textcolor{OliveGreen}{1505} & \textcolor{OliveGreen}{1535}   \\ \hline
Distribute-Three   & \textcolor{RawSienna}{0}    & \textcolor{RawSienna}{0}    & 0                             & \textcolor{OliveGreen}{2025}  & \textcolor{OliveGreen}{1510} & \textcolor{OliveGreen}{1487}   \\ \hline
Sum by Color       & \multicolumn{2}{r}{\textcolor{RawSienna}{0}}              & \multicolumn{4}{|r|}{\textcolor{OliveGreen}{24000}}                 \\ \hline
\end{tabular}
\caption{The Item Number for Each Abstract rule and Attribute in the Center Configuration of RAVEN.}
\label{tab:cs-raven-count}
\end{table*}

\begin{table*}[!h]
\centering
\begin{tabular}{ | c | c | c | c | c | c | c | }
\hline
                   & Number	                  & Position                    & Number/Position               & Type                          & Size                          & Color                         \\ \hline
Constant	       & 0                           & 0                           & \textcolor{OliveGreen}{970}   & \textcolor{OliveGreen}{1957}  & \textcolor{OliveGreen}{1527}  & \textcolor{OliveGreen}{1463}  \\ \hline
Progression	       & \textcolor{RawSienna}{556}  & \textcolor{RawSienna}{501}  & 0                             & \textcolor{OliveGreen}{1999}  & \textcolor{OliveGreen}{1512}  & \textcolor{OliveGreen}{1469}  \\ \hline
Arithmetic         & \textcolor{RawSienna}{989}  & \textcolor{RawSienna}{1038} & 0                             & 0                             & \textcolor{OliveGreen}{1503}  & \textcolor{OliveGreen}{1507}  \\ \hline
Distribute-Three   & \textcolor{RawSienna}{986}  & \textcolor{RawSienna}{960}  & 0                             & \textcolor{OliveGreen}{2044}  & \textcolor{OliveGreen}{1458}  & \textcolor{OliveGreen}{1561}  \\ \hline
Sum by Color       & \multicolumn{2}{r}{\textcolor{RawSienna}{5030}}           & \multicolumn{4}{|r|}{\textcolor{OliveGreen}{18970}}                  \\ \hline
\end{tabular}
\caption{The Item Number for Each Abstract rule and Attribute in the 2x2Grid Configuration of RAVEN.}
\label{tab:d4-raven-count}
\end{table*}

\begin{table*}[!h]
\centering
\begin{tabular}{ | c | c | c | c | c | c | c | }
\hline
                   & Number	                  & Position                    & Number/Position               & Type                          & Size                          & Color                         \\ \hline
Constant	       & 0                           & 0                           & \textcolor{OliveGreen}{833}   & \textcolor{OliveGreen}{1999}  & \textcolor{OliveGreen}{1494}  & \textcolor{OliveGreen}{1544}  \\ \hline
Progression	       & \textcolor{RawSienna}{828}  & \textcolor{RawSienna}{849}  & 0                             & \textcolor{OliveGreen}{1991}  & \textcolor{OliveGreen}{1552}  & \textcolor{OliveGreen}{1480}  \\ \hline
Arithmetic         & \textcolor{RawSienna}{872}  & \textcolor{RawSienna}{884}  & 0                             & 0                             & \textcolor{OliveGreen}{1553}  & \textcolor{OliveGreen}{1484}  \\ \hline
Distribute-Three   & \textcolor{RawSienna}{879}  & \textcolor{RawSienna}{855}  & 0                             & \textcolor{OliveGreen}{2010}  & \textcolor{OliveGreen}{1401}  & \textcolor{OliveGreen}{1492}  \\ \hline
Sum by Color       & \multicolumn{2}{r}{\textcolor{RawSienna}{5167}}           & \multicolumn{4}{|r|}{\textcolor{OliveGreen}{18833}}                            \\ \hline
\end{tabular}
\caption{The Item Number for Each Abstract rule and Attribute in the 3x3Grid Configuration of RAVEN.}
\label{tab:d9-raven-count}
\end{table*}

\begin{table*}[!h]
\centering
\begin{tabular}{ | c | c | c | c | c | c | c | }
\hline
                   & Number	                  & Position                    & Number/Position               & Type                          & Size                          & Color                         \\ \hline
Constant	       & 0                           & 0                           & \textcolor{OliveGreen}{6000}  & \textcolor{OliveGreen}{3378}  & \textcolor{OliveGreen}{3243} & \textcolor{OliveGreen}{6000}   \\ \hline
Progression	       & \textcolor{RawSienna}{0}    & \textcolor{RawSienna}{0}    & 0                             & \textcolor{OliveGreen}{3344}  & \textcolor{OliveGreen}{2412} & \textcolor{OliveGreen}{1511}   \\ \hline
Arithmetic         & \textcolor{RawSienna}{0}    & \textcolor{RawSienna}{0}    & 0                             & 0                             & \textcolor{OliveGreen}{1460} & \textcolor{OliveGreen}{1536}   \\ \hline
Distribute-Three   & \textcolor{RawSienna}{0}    & \textcolor{RawSienna}{0}    & 0                             & \textcolor{OliveGreen}{3320}  & \textcolor{OliveGreen}{3332} & \textcolor{OliveGreen}{1509}   \\ \hline
Sum by Color       & \multicolumn{2}{r}{\textcolor{RawSienna}{0}}              & \multicolumn{4}{|r|}{\textcolor{OliveGreen}{37045}}                                    \\ \hline
\end{tabular}
\caption{The Item Number for Each Abstract rule and Attribute in the Out-InCenter Configuration of RAVEN.}
\label{tab:cc-raven-count}
\end{table*}

\begin{table*}[!h]
\centering
\begin{tabular}{ | c | c | c | c | c | c | c | }
\hline
                   & Number	                  & Position                    & Number/Position               & Type                          & Size                          & Color                         \\ \hline
Constant	       & 0                           & 0                           & \textcolor{OliveGreen}{6000}  & \textcolor{OliveGreen}{3272}  & \textcolor{OliveGreen}{3432}  & \textcolor{OliveGreen}{6000}  \\ \hline
Progression	       & \textcolor{RawSienna}{473}  & \textcolor{RawSienna}{459}  & 0                             & \textcolor{OliveGreen}{3365}  & \textcolor{OliveGreen}{1894}  & \textcolor{OliveGreen}{1563}  \\ \hline
Arithmetic         & \textcolor{RawSienna}{1011} & \textcolor{RawSienna}{1011} & 0                             & 0                             & \textcolor{OliveGreen}{1707}  & \textcolor{OliveGreen}{1456}  \\ \hline
Distribute-Three   & \textcolor{RawSienna}{1060} & \textcolor{RawSienna}{976}  & 0                             & \textcolor{OliveGreen}{3385}  & \textcolor{OliveGreen}{3414}  & \textcolor{OliveGreen}{1490}  \\ \hline
Sum by Color       & \multicolumn{2}{r}{\textcolor{RawSienna}{4990}}           & \multicolumn{4}{|r|}{\textcolor{OliveGreen}{36978}}                                                                           \\ \hline
\end{tabular}
\caption{The Item Number for Each Abstract rule and Attribute in the Out-In2x2Grid Configuration of RAVEN.}
\label{tab:dc-raven-count}
\end{table*}

\begin{table*}[!h]
\centering
\begin{tabular}{ | c | c | c | c | c | c | c | }
\hline
                   & Number	                  & Position                    & Number/Position               & Type                          & Size                          & Color                         \\ \hline
Constant	       & 0                           & 0                           & \textcolor{OliveGreen}{6000}  & \textcolor{OliveGreen}{3392}  & \textcolor{OliveGreen}{2660} & \textcolor{OliveGreen}{2676}   \\ \hline
Progression	       & \textcolor{RawSienna}{0}    & \textcolor{RawSienna}{0}    & 0                             & \textcolor{OliveGreen}{3327}  & \textcolor{OliveGreen}{2623} & \textcolor{OliveGreen}{2617}   \\ \hline
Arithmetic         & \textcolor{RawSienna}{0}    & \textcolor{RawSienna}{0}    & 0                             & 0                             & \textcolor{OliveGreen}{2605} & \textcolor{OliveGreen}{2563}   \\ \hline
Distribute-Three   & \textcolor{RawSienna}{0}    & \textcolor{RawSienna}{0}    & 0                             & \textcolor{OliveGreen}{3378}  & \textcolor{OliveGreen}{2601} & \textcolor{OliveGreen}{2616}   \\ \hline
Sum by Color       & \multicolumn{2}{r}{\textcolor{RawSienna}{0}}              & \multicolumn{4}{|r|}{\textcolor{OliveGreen}{36958}}                                                                           \\ \hline
\end{tabular}
\caption{The Item Number for Each Abstract rule and Attribute in the Left-Right Configuration of RAVEN.}
\label{tab:lr-raven-count}
\end{table*}

\begin{table*}[!h]
\centering
\begin{tabular}{ | c | c | c | c | c | c | c | }
\hline
                   & Number	                  & Position                    & Number/Position               & Type                          & Size                          & Color                         \\ \hline
Constant	       & 0                           & 0                           & \textcolor{OliveGreen}{6000}  & \textcolor{OliveGreen}{3365}  & \textcolor{OliveGreen}{2575} & \textcolor{OliveGreen}{2610}   \\ \hline
Progression	       & \textcolor{RawSienna}{0}    & \textcolor{RawSienna}{0}    & 0                             & \textcolor{OliveGreen}{3362}  & \textcolor{OliveGreen}{2591} & \textcolor{OliveGreen}{2703}   \\ \hline
Arithmetic         & \textcolor{RawSienna}{0}    & \textcolor{RawSienna}{0}    & 0                             & 0                             & \textcolor{OliveGreen}{2679} & \textcolor{OliveGreen}{2605}   \\ \hline
Distribute-Three   & \textcolor{RawSienna}{0}    & \textcolor{RawSienna}{0}    & 0                             & \textcolor{OliveGreen}{3285}  & \textcolor{OliveGreen}{2639} & \textcolor{OliveGreen}{2612}   \\ \hline
Sum by Color       & \multicolumn{2}{r}{\textcolor{RawSienna}{0}}              & \multicolumn{4}{|r|}{\textcolor{OliveGreen}{37026}}                                                                           \\ \hline
\end{tabular}
\caption{The Item Number for Each Abstract rule and Attribute in the Up-Down Configuration of RAVEN.}
\label{tab:ud-raven-count}
\end{table*}

\begin{table*}[!h]
\centering
\begin{tabular}{ | c | c | c | c | c | c | c | }
\hline
                   & Number	                  & Position                    & Number/Position               & Type                          & Size                          & Color                         \\ \hline
Constant	       & 0                           & 0                           & \textcolor{OliveGreen}{1200}  & \textcolor{OliveGreen}{399}   & \textcolor{OliveGreen}{305}   & \textcolor{OliveGreen}{303}   \\ \hline
Progression	       & \textcolor{RawSienna}{0}    & \textcolor{RawSienna}{0}    & 0                             & \textcolor{OliveGreen}{416}   & \textcolor{OliveGreen}{277}   & \textcolor{OliveGreen}{293}   \\ \hline
Arithmetic         & \textcolor{RawSienna}{0}    & \textcolor{RawSienna}{0}    & 0                             & 0                             & \textcolor{OliveGreen}{306}   & \textcolor{OliveGreen}{295}   \\ \hline
Distribute-Three   & \textcolor{RawSienna}{0}    & \textcolor{RawSienna}{0}    & 0                             & \textcolor{OliveGreen}{385}   & \textcolor{OliveGreen}{312}   & \textcolor{OliveGreen}{309}   \\ \hline
Sum by Color       & \multicolumn{2}{r}{\textcolor{RawSienna}{0}}              & \multicolumn{4}{|r|}{\textcolor{OliveGreen}{4800}}                                                                           \\ \hline
\end{tabular}
\caption{The Item Number for Each Abstract rule and Attribute in the Center Configuration of AB-RAVEN.}
\label{tab:cs-ab-raven-count}
\end{table*}

\begin{table*}[!h]
\centering
\begin{tabular}{ | c | c | c | c | c | c | c | }
\hline
                   & Number	                  & Position                    & Number/Position               & Type                          & Size                          & Color                         \\ \hline
Constant	       & 0                           & 0                           & \textcolor{OliveGreen}{1277}  & \textcolor{OliveGreen}{4109}  & \textcolor{OliveGreen}{3193}  & \textcolor{OliveGreen}{3065}  \\ \hline
Progression	       & \textcolor{RawSienna}{1151} & \textcolor{RawSienna}{1124} & 0                             & \textcolor{OliveGreen}{4165}  & \textcolor{OliveGreen}{3147}  & \textcolor{OliveGreen}{3072}  \\ \hline
Arithmetic         & \textcolor{RawSienna}{2178} & \textcolor{RawSienna}{2253} & 0                             & 0                             & \textcolor{OliveGreen}{3127}  & \textcolor{OliveGreen}{3099}  \\ \hline
Distribute-Three   & \textcolor{RawSienna}{2149} & \textcolor{RawSienna}{2268} & 0                             & \textcolor{OliveGreen}{4126}  & \textcolor{OliveGreen}{3033}  & \textcolor{OliveGreen}{3164}  \\ \hline
Sum by Color       & \multicolumn{2}{r}{\textcolor{RawSienna}{11123}}          & \multicolumn{4}{|r|}{\textcolor{OliveGreen}{38477}}                                                                           \\ \hline
\end{tabular}
\caption{The Item Number for Each Abstract rule and Attribute in the 2x2Grid Configuration of AB-RAVEN.}
\label{tab:d4-ab-raven-count}
\end{table*}

\begin{table*}[!h]
\centering
\begin{tabular}{ | c | c | c | c | c | c | c | }
\hline
                   & Number	                  & Position                    & Number/Position               & Type                          & Size                          & Color                         \\ \hline
Constant	       & 0                           & 0                           & \textcolor{OliveGreen}{1097}  & \textcolor{OliveGreen}{4162}  & \textcolor{OliveGreen}{3133}  & \textcolor{OliveGreen}{3170}  \\ \hline
Progression	       & \textcolor{RawSienna}{1798} & \textcolor{RawSienna}{1850} & 0                             & \textcolor{OliveGreen}{4114}  & \textcolor{OliveGreen}{3126}  & \textcolor{OliveGreen}{3123}  \\ \hline
Arithmetic         & \textcolor{RawSienna}{1902} & \textcolor{RawSienna}{1912} & 0                             & 0                             & \textcolor{OliveGreen}{3212}  & \textcolor{OliveGreen}{3071}  \\ \hline
Distribute-Three   & \textcolor{RawSienna}{1916} & \textcolor{RawSienna}{1925} & 0                             & \textcolor{OliveGreen}{4124}  & \textcolor{OliveGreen}{2929}  & \textcolor{OliveGreen}{3036}  \\ \hline
Sum by Color       & \multicolumn{2}{r}{\textcolor{RawSienna}{11303}}            & \multicolumn{4}{|r|}{\textcolor{OliveGreen}{38297}}                                                                           \\ \hline
\end{tabular}
\caption{The Item Number for Each Abstract rule and Attribute in the 3x3Grid Configuration of AB-RAVEN.}
\label{tab:d9-ab-raven-count}
\end{table*}

\begin{table*}[!h]
\centering
\begin{tabular}{ | c | c | c | c | c | c | c | }
\hline
                   & Number	                  & Position                    & Number/Position               & Type                          & Size                          & Color                         \\ \hline
Constant	       & 0                           & 0                           & \textcolor{OliveGreen}{1200}  & \textcolor{OliveGreen}{650}   & \textcolor{OliveGreen}{653}  & \textcolor{OliveGreen}{1200}   \\ \hline
Progression	       & \textcolor{RawSienna}{0}    & \textcolor{RawSienna}{0}    & 0                             & \textcolor{OliveGreen}{697}   & \textcolor{OliveGreen}{495}  & \textcolor{OliveGreen}{295}   \\ \hline
Arithmetic         & \textcolor{RawSienna}{0}    & \textcolor{RawSienna}{0}    & 0                             & 0                             & \textcolor{OliveGreen}{325}  & \textcolor{OliveGreen}{315}   \\ \hline
Distribute-Three   & \textcolor{RawSienna}{0}    & \textcolor{RawSienna}{0}    & 0                             & \textcolor{OliveGreen}{669}   & \textcolor{OliveGreen}{640}  & \textcolor{OliveGreen}{313}   \\ \hline
Sum by Color       & \multicolumn{2}{r}{\textcolor{RawSienna}{0}}              & \multicolumn{4}{|r|}{\textcolor{OliveGreen}{7452}}                                                                           \\ \hline
\end{tabular}
\caption{The Item Number for Each Abstract rule and Attribute in the Out-InCenter Configuration of AB-RAVEN.}
\label{tab:cc-ab-raven-count}
\end{table*}

\begin{table*}[!h]
\centering
\begin{tabular}{ | c | c | c | c | c | c | c | }
\hline
                   & Number	                  & Position                    & Number/Position               & Type                          & Size                          & Color                         \\ \hline
Constant	       & 0                           & 0                           & \textcolor{OliveGreen}{12400} & \textcolor{OliveGreen}{6818}  & \textcolor{OliveGreen}{7040}  & \textcolor{OliveGreen}{12400}  \\ \hline
Progression	       & \textcolor{RawSienna}{1109} & \textcolor{RawSienna}{1066} & 0                             & \textcolor{OliveGreen}{6929}  & \textcolor{OliveGreen}{3920}  & \textcolor{OliveGreen}{3190}  \\ \hline
Arithmetic         & \textcolor{RawSienna}{2249} & \textcolor{RawSienna}{2142} & 0                             & 0                             & \textcolor{OliveGreen}{3518}  & \textcolor{OliveGreen}{3084}  \\ \hline
Distribute-Three   & \textcolor{RawSienna}{2270} & \textcolor{RawSienna}{2228} & 0                             & \textcolor{OliveGreen}{6950}  & \textcolor{OliveGreen}{7082}  & \textcolor{OliveGreen}{3050}  \\ \hline
Sum by Color       & \multicolumn{2}{r}{\textcolor{RawSienna}{11064}}         & \multicolumn{4}{|r|}{\textcolor{OliveGreen}{76381}}                                                                           \\ \hline
\end{tabular}
\caption{The Item Number for Each Abstract rule and Attribute in the Out-In2x2Grid Configuration of AB-RAVEN.}
\label{tab:dc-ab-raven-count}
\end{table*}

\begin{table*}[!h]
\centering
\begin{tabular}{ | c | c | c | c | c | c | c | }
\hline
                   & Number	                  & Position                    & Number/Position               & Type                          & Size                          & Color                         \\ \hline
Constant	       & 0                           & 0                           & \textcolor{OliveGreen}{1200}  & \textcolor{OliveGreen}{671}   & \textcolor{OliveGreen}{554}   & \textcolor{OliveGreen}{538}   \\ \hline
Progression	       & \textcolor{RawSienna}{0}    & \textcolor{RawSienna}{0}    & 0                             & \textcolor{OliveGreen}{660}   & \textcolor{OliveGreen}{522}   & \textcolor{OliveGreen}{519}   \\ \hline
Arithmetic         & \textcolor{RawSienna}{0}    & \textcolor{RawSienna}{0}    & 0                             & 0                             & \textcolor{OliveGreen}{495}   & \textcolor{OliveGreen}{537}   \\ \hline
Distribute-Three   & \textcolor{RawSienna}{0}    & \textcolor{RawSienna}{0}    & 0                             & \textcolor{OliveGreen}{671}   & \textcolor{OliveGreen}{540}   & \textcolor{OliveGreen}{515}   \\ \hline
Sum by Color       & \multicolumn{2}{r}{\textcolor{RawSienna}{0}}              & \multicolumn{4}{|r|}{\textcolor{OliveGreen}{7422}}                                                                           \\ \hline
\end{tabular}
\caption{The Item Number for Each Abstract rule and Attribute in the Left-Right Configuration of AB-RAVEN.}
\label{tab:lr-ab-raven-count}
\end{table*}

\begin{table*}[!h]
\centering
\begin{tabular}{ | c | c | c | c | c | c | c | }
\hline
                   & Number	                  & Position                    & Number/Position               & Type                          & Size                          & Color                         \\ \hline
Constant	       & 0                           & 0                           & \textcolor{OliveGreen}{1200}  & \textcolor{OliveGreen}{698}   & \textcolor{OliveGreen}{485}   & \textcolor{OliveGreen}{544}   \\ \hline
Progression	       & \textcolor{RawSienna}{0}    & \textcolor{RawSienna}{0}    & 0                             & \textcolor{OliveGreen}{690}   & \textcolor{OliveGreen}{511}   & \textcolor{OliveGreen}{518}   \\ \hline
Arithmetic         & \textcolor{RawSienna}{0}    & \textcolor{RawSienna}{0}    & 0                             & 0                             & \textcolor{OliveGreen}{525}   & \textcolor{OliveGreen}{531}   \\ \hline
Distribute-Three   & \textcolor{RawSienna}{0}    & \textcolor{RawSienna}{0}    & 0                             & \textcolor{OliveGreen}{654}   & \textcolor{OliveGreen}{544}   & \textcolor{OliveGreen}{520}   \\ \hline
Sum by Color       & \multicolumn{2}{r}{\textcolor{RawSienna}{0}}              & \multicolumn{4}{|r|}{\textcolor{OliveGreen}{7390}}                                                                           \\ \hline
\end{tabular}
\caption{The Item Number for Each Abstract rule and Attribute in the Up-Down Configuration of AB-RAVEN.}
\label{tab:ud-ab-raven-count}
\end{table*}

AB-RAVEN is a new variant of RAVEN. It is a more balanced dataset in terms of abstract concepts (i.e., the abstract rules and attribute used to generate matrix items) than the original RAVEN (and its variants I-RAVEN and RAVEN-FAIR). AB-RAVEN was created by increasing the number of training items of grid configurations and decreasing the number of training items of non-grid configurations, because the abstract concepts that are specific to grid configurations are much less represented in the original RAVEN (See Table 1 and 2 in the main paper). The increased part was generated using the original RAVEN generation code\footnote{https://github.com/WellyZhang/RAVEN}. The number of items of each configuration in AB-RAVEN is given in Tabel~\ref{tab:config-raven-ab-raven}. The AB-RAVEN dataset is available here\footnote{https://github.com/Yang-Yuan/CPCNet}.

The overall distribution of items over the abstract concepts in RAVEN (I-RAVEN and RAVEN-FAIR) and AB-RAVEN can be found in Table 1 and Table 2 in the main paper. Here, we provide the configuration-wise distribution of items over the abstract concepts in RAVEN (I-RAVEN and RAVEN-FAIR) and AB-RAVEN in Table~\ref{tab:cs-raven-count} through Table~\ref{tab:ud-ab-raven-count}.


\textcolor{red}{\noindent\textbf{WARNING:}}
\textbf{It needs to be pointed out that since we generated the new items of AB-RAVEN using the original RAVEN source code, AB-RAVEN has the same backdoor solution as the original RAVEN, i.e., comparing (embeddings of)  answer choices before scoring them leads to a high accuracy. However, as discussed in the experiment section of the main paper, this so-called backdoor is a double-edged sword. One one hand, it gives a shortcut (i.e., using multi-choice evaluation protocol) to achieve a high accuracy, which, however, does not really say anything about the abstract reasoning ability of the model being tested; on the other hand, combining the single-choice evaluation protocol with AB-RAVEN and RAVEN provides a more challenging and reasonable test to evaluate a model's abstract reasoning ability than using multi-choice evaluation protocol and/or I-RAVEN and/or RAVEN-FAIR. Thus, we encourage other researchers to use the positive edge of this sword to evaluate their models and restrain the idea of using the negative edge to just achieve high accuracies.}
\newpage

\section{CPCNet Implementation}
In all experiments in this paper, except the ablation studies (discussed in the next section), the $K$ and $L$ of the CPCNet model are set to 64 and 5, respectively. We used Tensorflow to implement our model. All the tensor data uses the data format of ``channel\_last" to boost running speed. For parameters not specified in this paper, they all take default values in Tensorflow 2.12.0. The source code can be found here\footnote{https://github.com/Yang-Yuan/CPCNet}. In the following of this section, we provide the implementation of Equation 1 through Equation 10 in the main paper. 

$F_E$ is a convolution-based entry encoder as shown in Table~\ref{tab:fe}.
\begin{table*}[!h]
    \centering
    \begin{tabular}{| l | l |} \hline
    Module Name & Specification \\ \hline 
    Conv2D      & channels = 32, kernel = 7, stride = 2, padding = ``same'', use\_bias = False \\ \hline
    BatchNormalization & axis = -1, momentum = 0.9 epsilon = 1e-5 \\ \hline
    Relu & - \\ \hline
    MaxPool2D & pool\_size = 3, stride = 2, padding = ``same'' \\ \hline
    Conv2D & channels = 64, kernel = 3, stride = 1, padding = ``same'', use\_bias = False \\ \hline
    BatchNormalization & axis = -1, momentum = 0.9 epsilon = 1e-5 \\ \hline
    Relu & - \\ \hline
    MaxPool2D & pool\_size = 3, stride = 2, padding = ``same'' \\ \hline
    \end{tabular}
    \caption{The implementation of the entry encoder $F_E$.}
    \label{tab:fe}
\end{table*}

For each $i \in \{1, 2, 3, 4, 5 \}$, $h^{(i)}_1, g^{(i)}_1, h^{(i)}_2$ and, $g^{(i)}_2$ follow the same residual structure as shown in Table~\ref{tab:res}. Since the tensor shape is not changed in the module, the residual link is implemented by directly adding the input to the tensor before the last ReLU.
\begin{table*}[!h]
    \centering
    \begin{tabular}{| l | l |} \hline
    Module Name & Specification \\ \hline 
    Conv2D      & channels = 64, kernel = 3, stride = 1, padding = ``same'', use\_bias = False \\ \hline
    BatchNormalization & axis = -1, momentum = 0.9 epsilon = 1e-5 \\ \hline
    Relu & - \\ \hline
    Conv2D & channels = 64, kernel = 3, stride = 1, padding = ``same'', use\_bias = False \\ \hline
    BatchNormalization & axis = -1, momentum = 0.9 epsilon = 1e-5 \\ \hline
    Relu & - \\ \hline
    \end{tabular}
    \caption{The implementation of the perceptual and conceptual processing $h^{(i)}_1, g^{(i)}_1, h^{(i)}_2$ and, $g^{(i)}_2$.}
    \label{tab:res}
\end{table*}

For each $i \in \{ 1, 2, 3, 4, 5 \}$, $q^{(i)}$ is a two-layer MLP as shown in Table~\ref{tab:con}.
\begin{table*}[!h]
    \centering
    \begin{tabular}{| l | l |} \hline
    Module Name & Specification \\ \hline 
    Dense       & units = 64, activation = ``relu", use\_bias = True \\ \hline
    Dense       & units = 64, activation = None, use\_bias = True \\ \hline
    \end{tabular}
    \caption{The implementation of the consistency computation $q^{(i)}$.}
    \label{tab:con}
\end{table*}

For each $j \in \{ 1 , 2 \} $, $p_j$ is a two-layer MLP as shown in Table~\ref{tab:mlp}.
\begin{table*}[!h]
    \centering
    \begin{tabular}{| l | l |} \hline
    Module Name & Specification \\ \hline 
    Dense       & units = 128, activation = ``relu", use\_bias = True \\ \hline
    Dense       & units = 1, activation = None, use\_bias = True \\ \hline
    \end{tabular}
    \caption{The implementation of the classification head $p_j$.}
    \label{tab:mlp}
\end{table*}

\newpage

\section{Experiments on I-RAVEN and RAVEN-FAIR}

\begin{table*}[h]
\centering
\begin{tabular}{l | l l l l l l l l l }
\toprule
& Model        & Avg. Acc.     & Center         & 2x2Grid        & 3x3Grid         & L-R             & U-D            & O-IC              & O-IG \\
\midrule
\multirow{2}{2.2cm}{Multi-Choice Evaluation Protocol}
&SRAN         & 60.8\%        & 78.2\%         & 50.1\%         & 42.4\%          & 70.1\%          & 70.3\%         & 68.2\%            & 46.3\%  \\
&SAVIR-T      & 98.1\%        & 99.5\%         & \textbf{98.1\%}& 93.8\%          & 99.6\%          & 99.1\%         & 99.5\%            & 97.2\%  \\
& & & & & & & & & \\ \hline
\multirow{7}{2.2cm}{Single-Choice Evaluation Protocol}
&NCD          & 48.22\%       & 60.00\%        & 31.20\%        & 29.95\%         & 58.90\%         & 57.15\%        & 62.35\%           & 39.00\% \\
&PrAE         & 77.02\%       & 90.45\%        & 85.35\%        & 45.60\%         & 96.25\%         & 97.35\%        & 63.45\%           & 60.70\% \\
&ALANS        & 78.5\%        & 72.3\%         & 79.5\%         & 72.9\%          & 79.2\%          & 79.6\%         & 85.9\%            & 79.9\%  \\
&NVSA         & 88.1\%        & 99.8\%         & 96.2\&         & 54.3\&          & 100.0\%         & 99.9\%         & 99.6\%            & 67.1\%  \\
&AlgebraMR    & 93.2\%        & 99.5\%         & 89.6\%         & 89.7\%          & 99.7\%          & 99.5\%         & 99.6\%            & 74.7\%  \\
&SCL          & 95.0\%        & 99.0\%         & 96.2\%         & 89.5\%          & 97.9\%          & 97.1\%         & 97.6\%            & 87.7\%  \\
&CPCNet(ours)&\textbf{98.5\%}&\textbf{100.0\%}&98.00\%    &\textbf{93.95\%}  &\textbf{100.0\%}  &\textbf{100.0\%}  & \textbf{100.0\%} &\textbf{97.55\%}\\ 
\bottomrule
\end{tabular}
\caption{Accuracies on the I-RAVEN. We report without-auxiliary-training accuracies of each model if possible. Data source for each row: \citep{hu2021stratified}, \citep{sahu2023savir}, \citep{zhuo2021unsupervised}, \citep{zhang2021abstract}, \citep{zhang2022learning}, \citep{hersche2023neuro}, \citep{xu2023abstract}, \citep{wu2021scattering}.}
\label{tab:acc-i-raven}
\end{table*}

\begin{table*}[h]
\centering
\begin{tabular}{l | l l l l l l l l l }
\toprule
        & Model    & Avg. Acc.     & Center         & 2x2Grid        & 3x3Grid         & L-R            & U-D           & O-IC           & O-IG \\
\midrule
\multirow{2}{2.2cm}{Multi-Choice Evaluation Protocol}
& SAVIR-T       & 97.4\%        & -              & -              & -               & -              & -             & -              & -        \\
& & & & & & & & & \\
& & & & & & & & & \\ \hline

\multirow{2}{2.2cm}{Single-Choice Evaluation Protocol}
& MRNet         & 86.8\%        & 97.0\%         & 72.7\%         & 69.5\%          & 98.7\%         & 98.9\%        & 97.6\%         & 73.3\%   \\
& CPCNet(ours)  & \textbf{98.14\%}& \textbf{100.0\%} & \textbf{98.15\%} & \textbf{90.80\%} & \textbf{100.0\%} & \textbf{100.0\%} & \textbf{100.0\%} & \textbf{98.00}\%\\
& & & & & & & & & \\ 
\bottomrule
\end{tabular}
\caption{Accuracies on RAVEN-FAIR. We report without-auxiliary-training accuracies of each model if possible. Data source for each row: \citep{sahu2023savir}, \citep{benny2021scale}.}
\label{tab:acc-raven-fair}
\end{table*}

All the experiments were done with tensorflow (python).

For the experiments to run fast, we made two choices for our experiment setting:
\begin{itemize}
    \item For some technical reasons (probably about the acceleration libraries that tensorflow uses), setting the random seed would reduce the training speed in our experiments. Thus, we did not set random seeds. Nonetheless, according to our experience, the results can be easily reproduced with default initializer of tensorflow and the hyper-parameters we provide in this supplementary material.
    \item Another key point is that we resized the matrix entry images from 160x160 to 80x80 to save some computation.
\end{itemize}
These two choices make it possible to run our experiments of RAVEN and its variants on a single consumer GPU (RTX 4090) and each trial took about 1 day to run.

Other hyper-parameters include:
\begin{itemize}
    \item Batch size of 32.
    \item Adam Optimizer with a starting learning rate of 0.0025, which is gradually increased to 0.2$\sim$0.4 and decreased to 0.0025. The purpose of doing so it to skip the local optima. Other advanced learning rate schedulers or optimizers can be used for the same purpose, but it is beyond the scope of this work.
\end{itemize}

For all the experiments, we monitored the validation accuracy. When it plateaued for several epochs, we took the checkpoints of the best validation accuracies for testing and reported the best test accuracy. In the main paper, we already provided the results of our model on the RAVEN and AB-RAVEN datasets. Here, we complement them with the results of our model on I-RAVEN and RAVEN-FAIR, as shown in Table~\ref{tab:acc-i-raven} and \ref{tab:acc-raven-fair}.

Note that the average accuracies in Table~\ref{tab:acc-i-raven} and \ref{tab:acc-raven-fair} are higher than the average accuracy of our model in Table~3. This again supported our argument that when using single-choice evaluation, i.e., not allowing the model to comparing answer choices before scoring them and thus not allowing it to use the backdoor of the original RAVEN, the original RAVEN is more challenging than I-RAVEN and RAVEN-FAIR, that is, the same model always achieves a higher accuracy on I-RAVEN and RAVEN-FAIR than on RAVEN. 
\newpage

\section{Ablation Studies on RAVEN}

\subsection{Varying the Number of Iterations $L$}
In Section 5 of the main paper, we denote the number of iterations of perceptual and conceptual processing by $L$. Here, we vary $L$ from 0 to 5, where 0 means no perceptual and conceptual processing at all and 5 is the number we used in the experiment in the main paper. Table~\ref{tab:ablation-raven-layers} shows the result of this ablation experiments. As $L$ increases from 0 to 5, the accuracy increases from 12.99\% to 96.92\%\footnote{Setting L greater than 5 could potentially further increase the accuracy but it is beyond our GPU memory limit.}. This suggests that stacking the layers of perceptual and conceptual processing significantly contributes to CPCNet's accuracy. Meanwhile, the increment gradually plateaus when $L \ge 3$ probably because the limited size of dataset dominates the performance of CPCNet. 

\begin{table*}[h]
\centering
\begin{tabular}{ l l l l l l l l l }
\toprule
      Model           & Avg. Acc.     & Center         & 2x2Grid        & 3x3Grid         & L-R             & U-D            & O-IC         & O-IG \\
\midrule
     CPCNet ($L$=0)   & 12.99\%       & 11.35\%        & 19.95\%        & 24.75\%         & 5.85\%          & 5.05\%         & 9.30\%       & 14.70\%  \\ 
     CPCNet ($L$=1)   & 77.32\%       & 87.05\%        & 56.95\%        & 57.20\%         & 94.50\%         & 95.35\%        & 93.55\%      & 56.65\%  \\ 
     CPCNet ($L$=2)   & 92.68\%       & 100.0\%        & 86.50\%        & 79.05\%         & 100.0\%         & 99.85\%        & 99.90\%      & 83.45\%  \\ 
     CPCNet ($L$=3)   & 94.22\%       & 100.0\%        & 88.30\%        & 79.55\%         & 100.0\%         & 99.90\%        & 99.90\%      & 91.90\%  \\ 
     CPCNet ($L$=4)   & 95.55\%       & 100.0\%        & 95.35\%        & 85.45\%         & 99.95\%         & 99.90\%        & 99.85\%      & 88.35\%  \\ 
     CPCNet ($L$=5)   & 96.92\%       & 100.0\%        & 96.70\%        & 86.05\%         & 100.0\%         & 99.90\%        & 99.90\%      & 95.90\%\\ 
\bottomrule
\end{tabular}
\caption{Ablation Accuracies of CPCNet on RAVEN by Varying $L$.}
\label{tab:ablation-raven-layers}
\end{table*}

\subsection{Ablating Consistency Computation}
As indicated in the main paper, CPCNet heavily relies on the idea of consistency between perceptual and conceptual processing and we particularly designed the two-path structure and special computation to achieve the consistency. Taking the CPCNet ($L$=5), the best-performing model in Table~\ref{tab:ablation-raven-layers}, we ablate these designs in different ways (see Figure~4 in the main paper):
\begin{enumerate}
    \item Removing the whole upper path in Figure~4 makes it more difficult for CPCNet to achieve the consistency between perceptual and conceptual processing. We thus name this kind of ablated CPCNet as \textbf{CPCNet-UP}. But note that ``difficult'' does not mean impossible as there are still both perceptual and conceptual processing in the lower path. For completely removing the possibility of achieving the consistency, CPCNet in Table~\ref{tab:ablation-raven-layers} ($L$=0) is an example. Similarly, we can remove the whole lower path and name the resulted model as \textbf{CPCNet-LP}.
    \item If we want to keep the two paths, we can still weaken the consistency by removing the internal consistency computation (i.e., the red arrows in Figure 4). We call this resulted ablated model \textbf{CPCNet-IC}.
    \item As noted in the main paper, the internal consistency computation just makes the consistency more likely to happen but does not guarantee that. We thus designed two  classification heads to push the conceptual and perceptual representations toward a consistent position by letting them classifying the same input. Therefore, we can have two more ablated model---\textbf{CPCNet-UC} and \textbf{CPCNet-LC}---by removing the upper and lower classification heads, respectively. Note that the input to the classification head is discarded when it is ablated.
\end{enumerate}

\begin{table*}[h]
\centering
\begin{tabular}{ l l l l l l l l l }
\toprule
Model       & Avg. Acc.     & Center         & 2x2Grid        & 3x3Grid         & L-R             & U-D            & O-IC         & O-IG     \\
\midrule
CPCNet-UP   & 93.05\%       & 99.75\%        & 87.95\%        & 80.15\%         & 100.0\%         & 99.80\%        & 99.80\%      & 83.90\%  \\ 
CPCNet-LP   & 91.01\%       & 99.90\%        & 79.65\%        & 75.45\%         & 99.95\%         & 99.85\%        & 99.80\%      & 82.45\%  \\ 
CPCNet-IC   & 94.19\%       & 100.0\%        & 90.45\%        & 81.65\%         & 99.95\%         & 99.90\%        & 99.95\%      & 87.40\%  \\ 
CPCNet-UC   & 96.69\%       & 100.0\%        & 95.90\%        & 89.15\%         & 100.0\%         & 99.80\%        & 99.90\%      & 92.05\%  \\ 
CPCNet-LC   & 95.11\%       & 100.0\%        & 93.85\%        & 84.95\%         & 100.0\%         & 99.85\%        & 99.95\%      & 87.15\%  \\ \hline
CPCNet ($L$=5) & 96.92\%       & 100.0\%        & 96.70\%        & 86.05\%         & 100.0\%         & 99.90\%        & 99.90\%      & 95.90\%\\ 
CPCNet-original & 96.29\%       & 100.0\%        & 94.50\%        & 86.40\%         & 99.90\%         & 99.85\%        & 99.90\%      & 93.45 \\
\bottomrule
\end{tabular}
\caption{Ablation Accuracies of CPCNet on RAVEN by Weakening Consistency in Different Ways.}
\label{tab:ablation-raven-consistency}
\end{table*}

Table~\ref{tab:ablation-raven-consistency} shows the results of ablating consistency design in different ways. In the last row, we have a model named CPCNet-original to tell a interesting fact that when we trained the full CPCNet ($L$=5) for the first time, its test accuracy was 96.29\%, as shown in the table. It is interesting because the accuracy is lower than the accuracy 96.69\% of its ablated version CPCNet-UC. This is very counter-intuitive because the ablated part should be beneficial for learning according to its design purpose. Thus, we retrained the model for longer and adjusted learning rate so that it could skip local optima (if any). This gave us the accuracy 96.92\% of CPCNet ($L$=5), which is greater than CPCNet-UC's and seems a reasonable result. But we did not retrained the ablated models and it is unclear whether or not they could have higher accuracies and that the ablated ones are actually more capable models.

\section{Experiments on PGM}
We also tested CPCNet on another important benchmark dataset---PGM \citep{barrett2018measuring}. Note that in this work, we only tested the neutral generalization regime of PGM, which is considered as a basic usage of PGM. However, it needs to be pointed out that the other non-neutral generalization regimes of PGM are extremely important, representing a different type of generalization tasks, which is beyond the scope of this paper. See our experimental results on PGM in Table~\ref{tab:acc-pgm-neutral}.

\begin{table*}[h]
\centering
\begin{tabular}{l | l l}
\toprule
& Model        & Avg. Acc.  \\
\midrule
\multirow{3}{2.2cm}{Multi-Choice Evaluation Protocol}
&CoPINet        & 56.37\%      \\ 
&MXGNet         & 66.7\%       \\
&LEN            & 68.1\%       \\
&DCNet          & 68.57\%      \\
&SRAN           & 71.3\%       \\
&SAVIR-T        & 91.2\%       \\
& &  \\ \hline
\multirow{3}{2.2cm}{Single-Choice Evaluation Protocol}
&ARNe           & 12.55\% \\
&WReN           & 62.6\% \\
&Rel-Base       & 85.5\% \\
&SCL            & 88.9\% \\
&MRNet          & 93.4\% \\
&RS-TRAN        & 97.5\%       \\
&MLRN           & 98.03\%       \\
&CPCNet(ours)   &\textbf{98.4\%}\\ 
\bottomrule
\end{tabular}
\caption{Accuracies on the PGM-Neutral. We report without-auxiliary-training accuracies of each model if possible. Data source for each row: \citep{zhang2019learning}, \citep{wang2020abstract}, \citep{zheng2019abstract}, \citep{zhuo2021effective}, \citep{hu2021stratified}, \citep{sahu2023savir}, \citep{hahne2019attention}, \citep{barrett2018measuring}, \citep{spratley2020closer}, \citep{wu2021scattering}, \citep{benny2021scale}, \citep{wei2023multi}, \citep{jahrens2020solving}.}
\label{tab:acc-pgm-neutral}
\end{table*}



\end{document}